\pdfoutput=1

\documentclass[11pt]{article}

\usepackage{ACL2023}

\usepackage{mathptmx}
\usepackage{latexsym}
\usepackage{graphicx}
\usepackage{caption}
\usepackage{subcaption}
\usepackage{multirow}
\usepackage{latexsym}

\usepackage[T1]{fontenc}

\usepackage[utf8]{inputenc}

\usepackage{microtype}

\usepackage{inconsolata}
\usepackage{tikz}
\usetikzlibrary{positioning, shapes.geometric}
\title{Large Language Models on Fine-grained Emotion Detection Dataset with Data Augmentation and Transfer Learning}

\author{Kaipeng Wang \\
  School of Computer Science \\
  Carnegie Mellon University \\
  \texttt{kaipeng2@cs.cmu.edu} \\\And
  Zhi Jing \\
  School of Computer Science \\
  Carnegie Mellon University \\
  \texttt{zjing2@cs.cmu.edu} \\\AND
  Yongye Su \\
  Department of Computer Science \\
  Purdue University \\
  \texttt{su311@purdue.edu} \\\And
  Yikun Han \\
  Department of Statistics \\
  University of Michigan \\
  \texttt{yikunhan@umich.edu} \\}

\begin{document}
\maketitle

\begin{abstract}
This paper delves into enhancing the classification performance on the GoEmotions dataset, a large, manually annotated dataset for emotion detection in text. The primary goal of this paper is to address the challenges of detecting subtle emotions in text, a complex issue in Natural Language Processing (NLP) with significant practical applications. The findings offer valuable insights into addressing the challenges of emotion detection in text and suggest directions for future research, including the potential for a survey paper that synthesizes methods and performances across various datasets in this domain.
\end{abstract}

\section{Introduction}

Recently Large Language Models have gain enormous attention in the field of Natural Language Processing. The task of detecting subtle emotions in text is a complex issue in Natural Language Processing with numerous practical uses. A major hurdle is the lack of sufficiently large and annotated datasets for emotion detection. Demszky \cite{demszky-etal-2020-goemotions} created the GoEmotions dataset, which is the largest manually annotated dataset containing 58k English Reddit comments, labeled for 27 emotion categories and Neutral. The researchers also fine-tuned a BERT model on GoEmotions as a strong baseline in transfer learning across domains and taxonomies to show the potentials of this dataset.

In the initial experiments, our team dived deep into the original paper and reproduced the results of both fine-tuning BERT model on GoEmotions dataset and applying transfer learning technique using fine-tuned BERT to other emotion classification dataset with different taxonomies. We were able to achieve 0.49 macro-average F1 score on the 28-label classification task, which was slightly improved over the best result 0.46 reported in the original paper \cite{demszky-etal-2020-goemotions}. However, we believe that there is still much room for improvement in term of classification performance on GoEmotions dataset. Therefore, our main objective in this study is to utilize different methods to enhance the classification performance on the fine-grained emotion detection dataset, GoEmotions, as much as possible.

\section{Related Work}
Our study builds upon the foundation laid by GoEmotions \cite{demszky-etal-2020-goemotions}, a pioneering framework in emotion classification within textual content. GoEmotions excels in its comprehensive categorization, identifying expressions across 28 distinct emotional categories (27 emotions plus a neutral category). This framework is instrumental in enhancing sentiment analysis and interpreting a broad spectrum of human emotions, particularly in the context of concise social media posts.

In parallel, we draw insights from SODA (SOcial DiAlogues) \cite{kim2023soda}, a unique dataset in the realm of social dialogue systems. SODA synthesizes 1.5 million dialogues, integrating socially-grounded conversations from a pre-trained language model with contextual knowledge from the Atomic10x knowledge graph \cite{west2022symbolic}. This integration addresses key challenges in dialogue agent training, namely diversity, scale, and quality, thereby surpassing traditional human-authored corpora in terms of consistency, specificity, and naturalness. SODA's contributions are pivotal in developing dialogue agents that are more authentic and coherent.

Additionally, our work touches upon the critical concept of predictive uncertainty in Large Language Models (LLMs) \cite{jing2024large, su2024large}. This uncertainty bifurcates into aleatoric uncertainty, rooted in inherent randomness in data, and epistemic uncertainty, originating from incomplete data knowledge. Recent scholarly efforts have been directed towards addressing these uncertainties in deep neural models, highlighting their significance in enhancing model reliability.

A notable development in this domain is presented in \cite{li2023cue}, which introduces the CUE (Contextual Uncertainty Elimination) framework. This innovative approach leverages a Variational Auto-encoder to modify latent text representations, thereby elucidating the sources of predictive uncertainty in LLM classifiers. CUE extends beyond mere identification of influential input tokens; it recalibrates text representations to provide a novel lens for interpreting uncertainty in LLM-based text classifiers. This advancement is pivotal in our understanding of LLM behavior, particularly in enhancing model transparency and developing effective strategies for uncertainty mitigation.

Our research is situated at the intersection of these significant developments, aiming to further the understanding and application of emotion classification in LLMs, while addressing the challenges and uncertainties inherent in these sophisticated models.

\section{Limitations and Hypotheses} \label{limitations}
As previously mentioned, the GoEmotions dataset contains 58k Reddit comments, which are classified into 27 emotion categories plus neutral. During the annotation process, there are three annotators labeling the same data points and multiple labels for a data points are allowed. Although the creation of the dataset seems sound, this fine-grained emotion classification dataset still has limitations.

\begin{figure}[tp]
    \centering
    \includegraphics[width=0.45\textwidth]{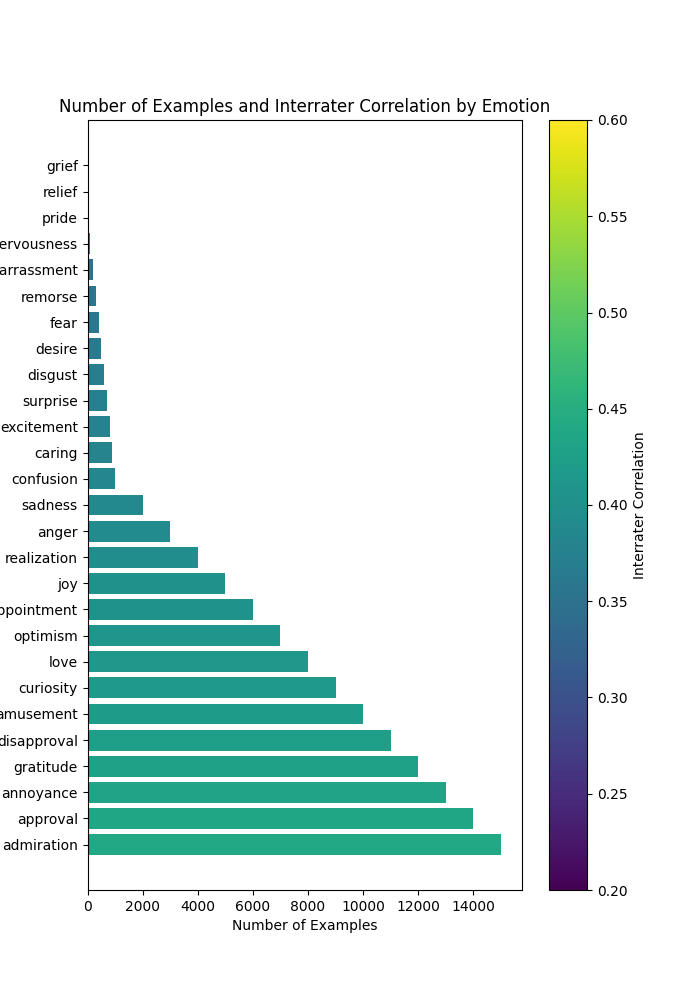}
    \caption{Emotion Categories Ordered by Number of Examples in the Dataset\cite{demszky-etal-2020-goemotions}}
    \label{fig:distribution}
\end{figure}

Firstly, the emotion categories in GoEmotions dataset have uneven distributions in nature. As shown in Figure \ref{fig:distribution}, some emotion categories such as "admiration", "approval", and "annoyance" have more than 10k examples in the dataset, while some minority classes such as "pride", "relief", and "grief" only have less than 1k examples. As a result, there will be an approximately 10 times difference in training data sample size between some majority classes and minority classes, causing a nonnegligible difference on single category classification performance. For instance, Demszky and his team fine-tuned a BERT model on GoEmotions dataset and the "grief" category which has the least sample size in training set achieved all zeros across different evaluation metrics \cite{demszky-etal-2020-goemotions}.

The second limitation of GoEmotions is that the dataset contains biases and is not representative of global diversity. Since the annotation process is all done by Indian English speakers and all data points come from English Reddit community, potential biases could be introduced during the data collecting and annotation process. To improve on this and remove biases, we may consider using annotators with more representative background and collecting data from multiple resources. These improvement methods are valuable for future work but may be out of this project's scope.

The third limitation results from the reliance on traditional language models, such as BERT, in the GoEmotions paper which was published four years ago. There is a lack of exploration into how modern large language models (LLMs) like GPT-4 and Llama would perform on emotion detection datasets and into their potential for understanding and interpreting human emotions.

Therefore, inspired by the first limitation of the dataset and several common practices we learned in class, our team has come up with three hypotheses to further improve the classification performance on GoEmotions dataset:
\begin{enumerate}
    \item State-of-the-art model like RoBERTa could outperform BERT on the GoEmotions dataset
    \item Data augmentation methods applied on GoEmotions dataset could improve the classification performance
    \item Transfer learning on a dataset with similar domain and different taxonomy could further improve the classification performance on GoEmotions dataset
    \item Modern LLMs like GPT-4 and Llama will, under zero-shot settings, outperform traditionally fine-tuned Language Models. This paper is the first to evaluate the capabilities of LLMs on the GoEmotions dataset.
\end{enumerate}

Based on the aforementioned hypotheses, we designed three sets of experiments to validate our hypotheses which are briefly listed below and discussed in detail in section \ref{exp}:
\begin{enumerate}
    \item Reproduce the experiments from GoEmotions Dataset paper \cite{demszky-etal-2020-goemotions} and compare the performance with the published benchmark. This step is to ensure our experiment environment are correctly set because it has been 4 years and the original paper trained the networks using TensorFlow framework whereas we are using HuggingFace and PyTorch.
    \item Fine-tune RoBERTa model on GoEmotions dataset and compare the performance with BERT benchmark.
    \item Apply three data augmentation techniques on original training dataset before fine-tuning RoBERTa, comparing the variance in performance.
    \item Fine-tune RoBERTa on CARER dataset \cite{saravia-etal-2018-carer} first and then fine-tune the trained model on augmented GoEmotions training dataset, comparing the variance in performance.
    \item Introduce modern LLMs like Llama \cite{touvron2023llama} and GPT-4 \cite{openai2024gpt4} and test them on a subset of GoEmotions and compare them with BERT benchmark. This paper is the first to put LLMs in test of GoEmotions dataset.
\end{enumerate}

\section{Experiments and Results} \label{exp}

\subsection{Fine-tuning BERT on GoEmotions}
The first set of experiments in the original GoEmotions paper \cite{demszky-etal-2020-goemotions} is to fine-tune a pre-trained bert-base-cased model on the GoEmotions dataset with three different taxonomy hierarchies. The first hierarchy we called it the original taxonomy, which contains 27 emotion labels and a neutral label. The second hierarchy is called the grouped taxonomy. The grouped taxonomy is generated by applying a hierarchical clustering on the original 27 labels using correlation as a distance metric, resulting in 3 categories and a neutral category. The last hierarchy is called the Ekman’s taxonomy \cite{ekman1992argument}, which maps the original 27 emotion labels into 6 groups with a neutral group.

Therefore, 3 taxonomies correspond to 3 different multi-label emotion classification problem. As described in the original paper, we fine-tune a bert-base-cased model on the GoEmotions dataset three times, one for each taxonomy. The model structure will be different only in the last linear layer to fit the number of classes to predict. We keep the exactly same hyperparameter configuration as the original paper: batch\_size of 16, learning\_rate of 5e-5 and no learning rate scheduler. We use the AdamW optimizer \cite{loshchilov2017decoupled} and cross-entropy as the loss function.

We trained 10 epochs for each classification task, and selected the best model using the macro-F1 score on the dev dataset. For the original, group, and Ekman’s taxonomy, the best dev macro-F1 are achieved at the $9^{th}$, $2^{nd}$, and $2^{nd}$ epoch respectively. Finally, we evaluated the performances of our best models on the test set. The detailed results on 3 taxonomies can be found in Table \ref{tab:bert-original}, Table \ref{tab:bert-group}, and Table \ref{tab:bert-ekman}.

\begin{table}[tp]
\centering
\begin{tabular}{lccc}
\hline
\textbf{Emotion} & \textbf{Precision} & \textbf{Recall} & \textbf{F1}\\
\hline
admiration & 0.62 & 0.73 & 0.67 \\
amusement & 0.72 & 0.88 & 0.79 \\
anger & 0.47 & 0.48 & 0.48 \\
annoyance & 0.31 & 0.40 & 0.35 \\
approval & 0.33 & 0.41 & 0.36 \\
caring & 0.39 & 0.45 & 0.42 \\
confusion & 0.39 & 0.48 & 0.43 \\
curiosity & 0.44 & 0.53 & 0.48 \\
desire & 0.50 & 0.41 & 0.45 \\
disappointment & 0.31 & 0.27 & 0.29 \\
disapproval & 0.37 & 0.40 & 0.38 \\
disgust & 0.55 & 0.46 & 0.50 \\
embarrassment & 0.44 & 0.43 & 0.44 \\
excitement & 0.43 & 0.45 & 0.44 \\
fear & 0.62 & 0.71 & 0.66 \\
gratitude & 0.90 & 0.92 & 0.91 \\
grief & 0.43 & 0.50 & 0.46 \\
joy & 0.54 & 0.67 & 0.60 \\
love & 0.75 & 0.84 & 0.80 \\
nervousness & 0.31 & 0.48 & 0.38 \\
optimism & 0.55 & 0.56 & 0.55 \\
pride & 0.83 & 0.31 & 0.45 \\
realization & 0.21 & 0.23 & 0.22 \\
relief & 0.29 & 0.36 & 0.32 \\
remorse & 0.56 & 0.77 & 0.65 \\
sadness & 0.57 & 0.60 & 0.58 \\
surprise & 0.52 & 0.57 & 0.55 \\
neutral & 0.61 & 0.72 & 0.66 \\
\textbf{macro-average} & \textbf{0.50} & \textbf{0.54} & \textbf{0.51} \\
\textbf{std} & \textbf{0.16} & \textbf{0.18} & \textbf{0.16} \\

\hline
\end{tabular}
\caption{\label{tab:bert-original}
Results based on GoEmotions taxonomy
}
\end{table}

\begin{table}[tp]
\centering
\begin{tabular}{lccc}
\hline
\textbf{Sentiment} & \textbf{Precision} & \textbf{Recall} & \textbf{F1}\\
\hline
ambiguous & 0.50 & 0.74 & 0.60 \\
negative & 0.56 & 0.81 & 0.66 \\
neutral & 0.63 & 0.72 & 0.67 \\
positive & 0.80 & 0.84 & 0.82 \\
\textbf{macro-average} & \textbf{0.62} & \textbf{0.78} & \textbf{0.69} \\
\textbf{std} & \textbf{0.11} & \textbf{0.05} & \textbf{0.08} \\
\hline
\end{tabular}
\caption{\label{tab:bert-group}
Results based on sentiment-grouped data
}
\end{table}

\begin{table}[tp]
\centering
\begin{tabular}{cccc}
\hline
\textbf{Model} & \textbf{Precision} & \textbf{Recall} & \textbf{F1}\\
\hline
BERT & 0.40 & \textbf{0.63} & 0.46 \\
RoBERTa & \textbf{0.50} & 0.54 & \textbf{0.51} \\
\hline
\end{tabular}
\caption{
compared models
}
\end{table}

\begin{table}[tp]
\centering
\begin{tabular}{cccc}
\hline
\textbf{Training} & \textbf{Precision} & \textbf{Recall} & \textbf{F1}\\
\hline
Original & 0.50 & 0.54 & 0.51 \\
DDA & IP & IP & IP \\
BERT Embed & IP & IP & IP \\
ProtAug & IP & IP & IP \\
\hline
\end{tabular}
\caption{
data augmentation
}
\end{table}

\begin{table}[tp]
\centering
\begin{tabular}{lccc}
\hline
\textbf{Ekman Emotion} & \textbf{Precision} & \textbf{Recall} & \textbf{F1}\\
\hline
anger & 0.56 & 0.52 & 0.54 \\
disgust & 0.44 & 0.54 & 0.49 \\
fear & 0.55 & 0.74 & 0.63 \\
joy & 0.73 & 0.89 & 0.80 \\
neutral & 0.59 & 0.80 & 0.68 \\
sadness & 0.54 & 0.63 & 0.58 \\
surprise & 0.53 & 0.70 & 0.60 \\
\textbf{macro-average} & \textbf{0.56} & \textbf{0.69} & \textbf{0.62} \\
\textbf{std} & \textbf{0.08} & \textbf{0.13} & \textbf{0.10} \\
\hline
\end{tabular}
\caption{\label{tab:bert-ekman}
Results using Ekman’s taxonomy
}
\end{table}

Our final results are comparable with the original paper. We list all numerical results for comparison (the numbers inside the brackets come from the original paper). We achieved macro-Precision of 0.50 (0.40), macro-Recall of 0.54 (0.63), and macro-F1 of 0.51 (0.46) on GoEmotions taxonomy; macro-Precision of 0.62 (0.65), macro-Recall of 0.78 (0.74), and macro-F1 of 0.69 (0.69) on grouped taxonomy; macro-Precision of 0.56 (0.59), macro-Recall of 0.69 (0.69), and macro-F1 of 0.62 (0.64) on Ekman’s taxonomy.

\begin{figure*}[h!]
  \centering
  \begin{subfigure}[b]{0.32\textwidth}
    \centering
    \includegraphics[width=\textwidth]{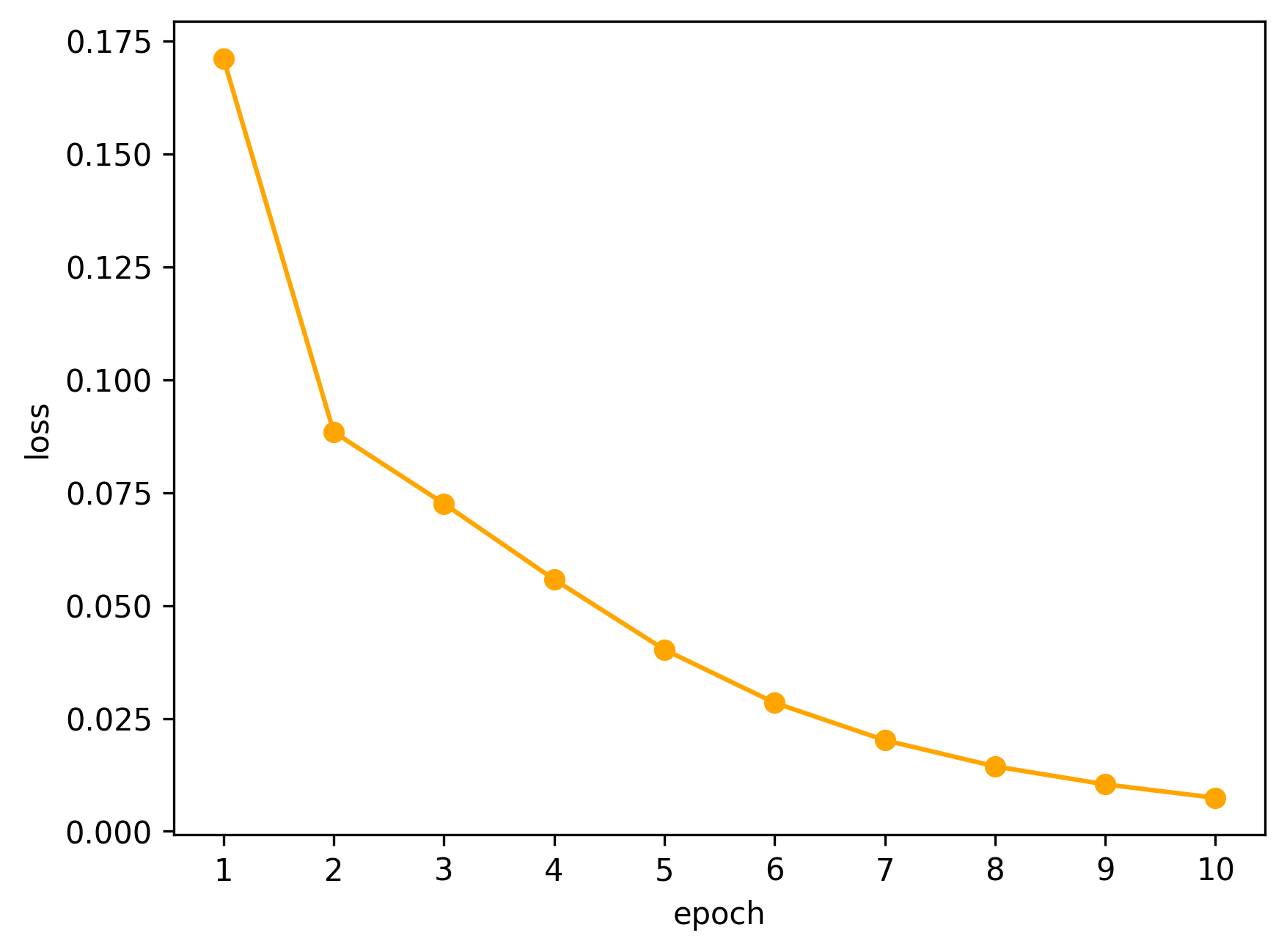}
    \caption{Losses on GoEmotions Taxonomy}
    \label{fig:losses_original}
  \end{subfigure}
  \hfill
  \begin{subfigure}[b]{0.32\textwidth}
    \centering
    \includegraphics[width=\textwidth]{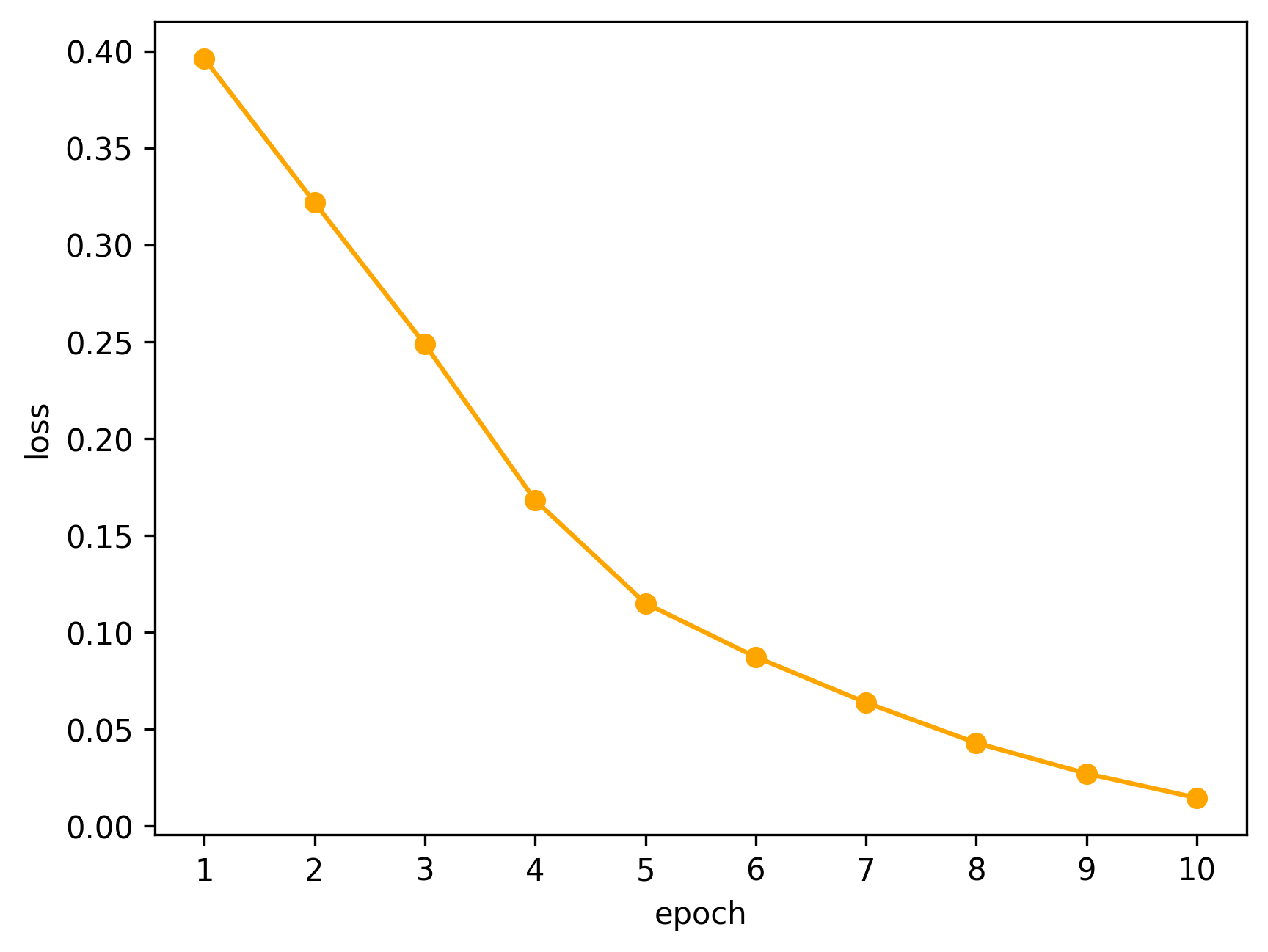}
    \caption{Losses on Grouped Taxonomy}
    \label{fig:losses_group}
  \end{subfigure}
  \hfill
  \begin{subfigure}[b]{0.32\textwidth}
    \centering
    \includegraphics[width=\textwidth]{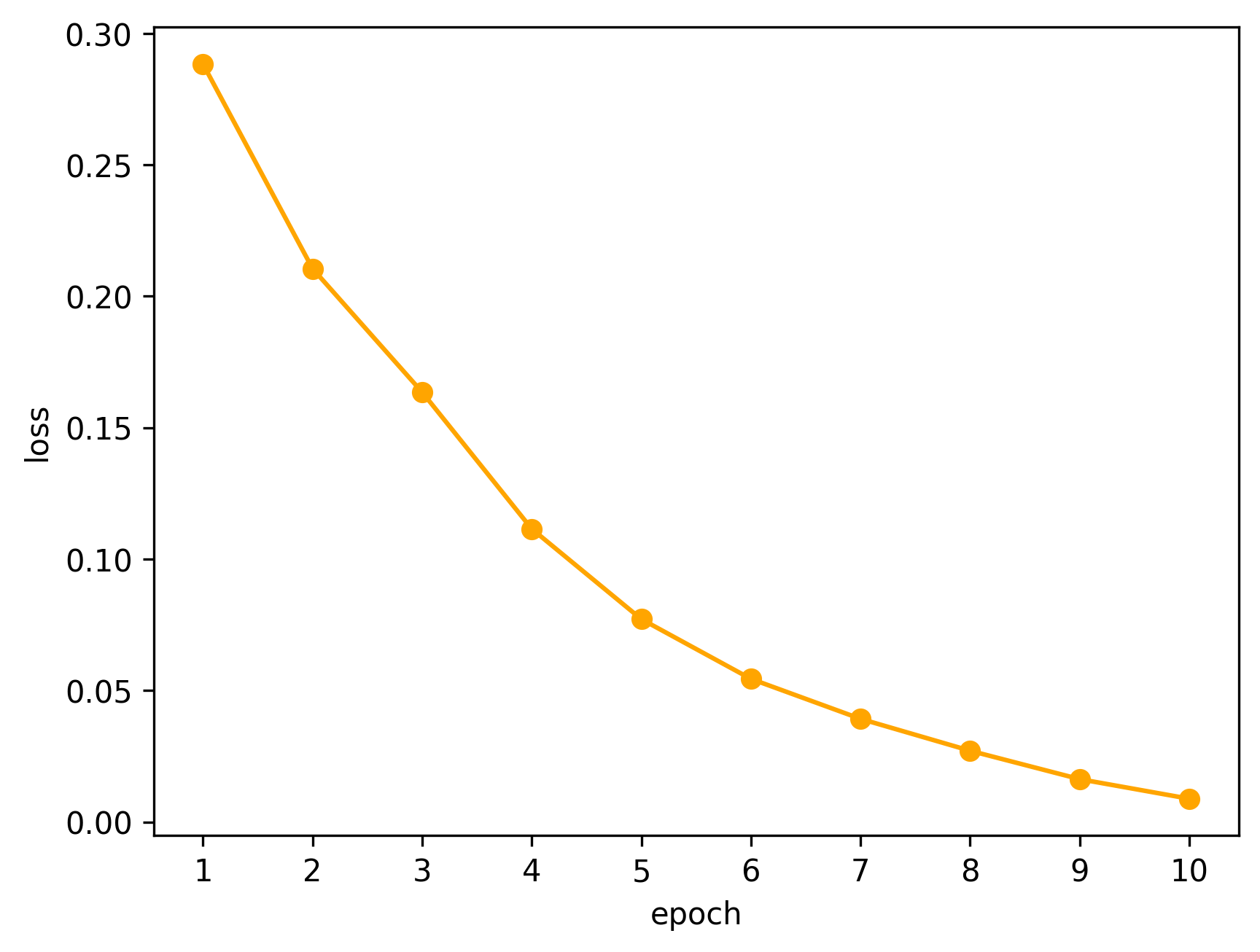}
    \caption{Losses on Ekman’s taxonomy}
    \label{fig:losses_ekman}
  \end{subfigure}

  \caption{Fine-tuning Losses on Three Taxonomies}
  \label{fig:losses}
\end{figure*}

We also plotted the training losses of three tasks in Figure \ref{fig:losses}. Figure \ref{fig:losses_original}, Figure \ref{fig:losses_group}, and Figure \ref{fig:losses_ekman} prove that the argument in the original paper stating that 4 epochs are enough to acquire strong performance when fine-tuning BERT on the GoEmotions dataset.

\subsection{Transfer Learning Experiment}

In the paper, the authors also conducted transfer learning experiments on existing emotion benchmarks.

The experiment aims to demonstrate the effectiveness of transfer learning in emotion understanding across various domains and taxonomies using limited labeled data in a target domain. The researchers leverage GoEmotions\cite{demszky-etal-2020-goemotions} as the baseline dataset for training their models.

The study uses nine datasets from the \cite{bostan-klinger-2018-analysis} Unified Dataset, focusing on three for detailed discussion due to their diverse domains. The chosen datasets are:

\begin{enumerate}
    \item ISEAR\cite{ISEAR}: This dataset includes personal reports of emotional events from 3000 individuals with different cultural backgrounds. It consists of 8,000 sentences, each labeled with one of seven emotions: anger, disgust, fear, guilt, joy, sadness, and shame.
    \item EmoInt\cite{marresetaylor2017emoatt}: Part of the SemEval 2018 benchmark, this dataset has 7,000 tweets with crowdsourced annotations. It focuses on the intensity of four emotions: anger, joy, sadness, and fear. The researchers use a cutoff of 0.5 to obtain binary annotations.
    \item Emotion-Stimulus\cite{mohammad-etal-2018-semeval}: Based on FrameNet's emotion-directed frames, this dataset provides annotations for 2,400 sentences. Its taxonomy includes seven emotions: anger, disgust, fear, joy, sadness, shame, and surprise.
\end{enumerate}

The experiments involve varying training set sizes, including 100, 200, 500, 1,000, and 80\% of the dataset examples. For each size, 10 random splits are created, with the remaining examples serving as a test set. The results are reported with confidence intervals based on repeated experiments. All experiments maintain a batch size of 16, learning rate of 2e-5, and three epochs.

The goal of these experiments is to establish that GoEmotions can serve as a robust baseline for understanding emotions, even with limited labeled data in different domains, and to assess the effectiveness of different finetuning approaches in this context.

In this assignment 3, we show our reproducing results on ISEAR mainly because the other 2 experiments are highly similar with ISEAR, and the entire experiment is not considered as the major one. We use the exact setting of the paper mentioned which is summarized above. The difference is we use the BERT finetuned on Ekman as our base model because we find the labels from Ekman and ISEAR take high resemblance of each other.

Below in Figure \ref{fig:TransferLearning} is the full our reproducing results on ISEAR.

\begin{figure*}[h!]
  \centering
  \begin{subfigure}[b]{0.32\textwidth}
    \centering
    \includegraphics[width=\textwidth]{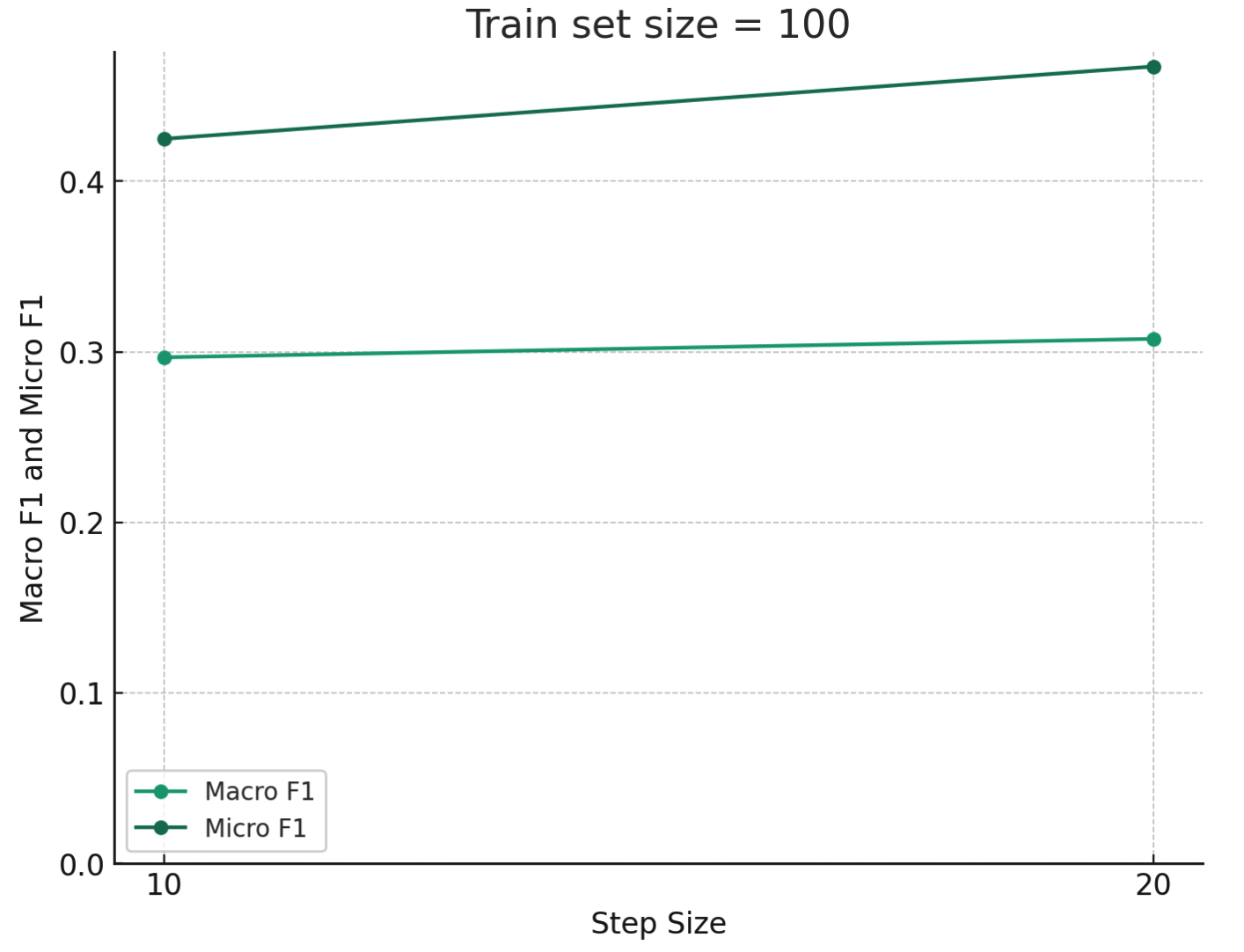}
    \caption{Train set size = 100}
    \label{fig:train100}
  \end{subfigure}
  \hfill
  \begin{subfigure}[b]{0.32\textwidth}
    \centering
    \includegraphics[width=\textwidth]{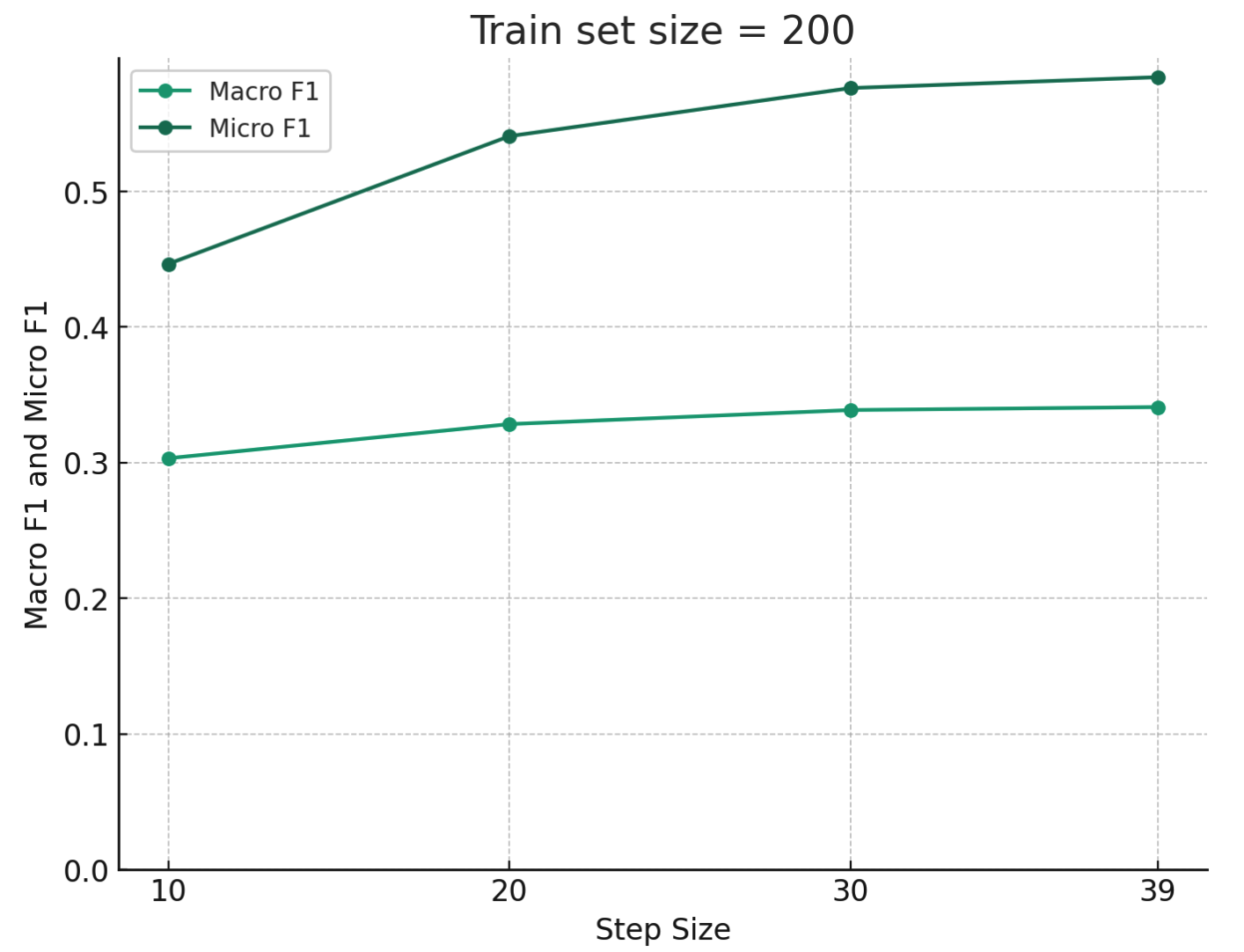}
    \caption{Train set size = 200}
    \label{fig:train200}
  \end{subfigure}
  \hfill
  \begin{subfigure}[b]{0.32\textwidth}
    \centering
    \includegraphics[width=\textwidth]{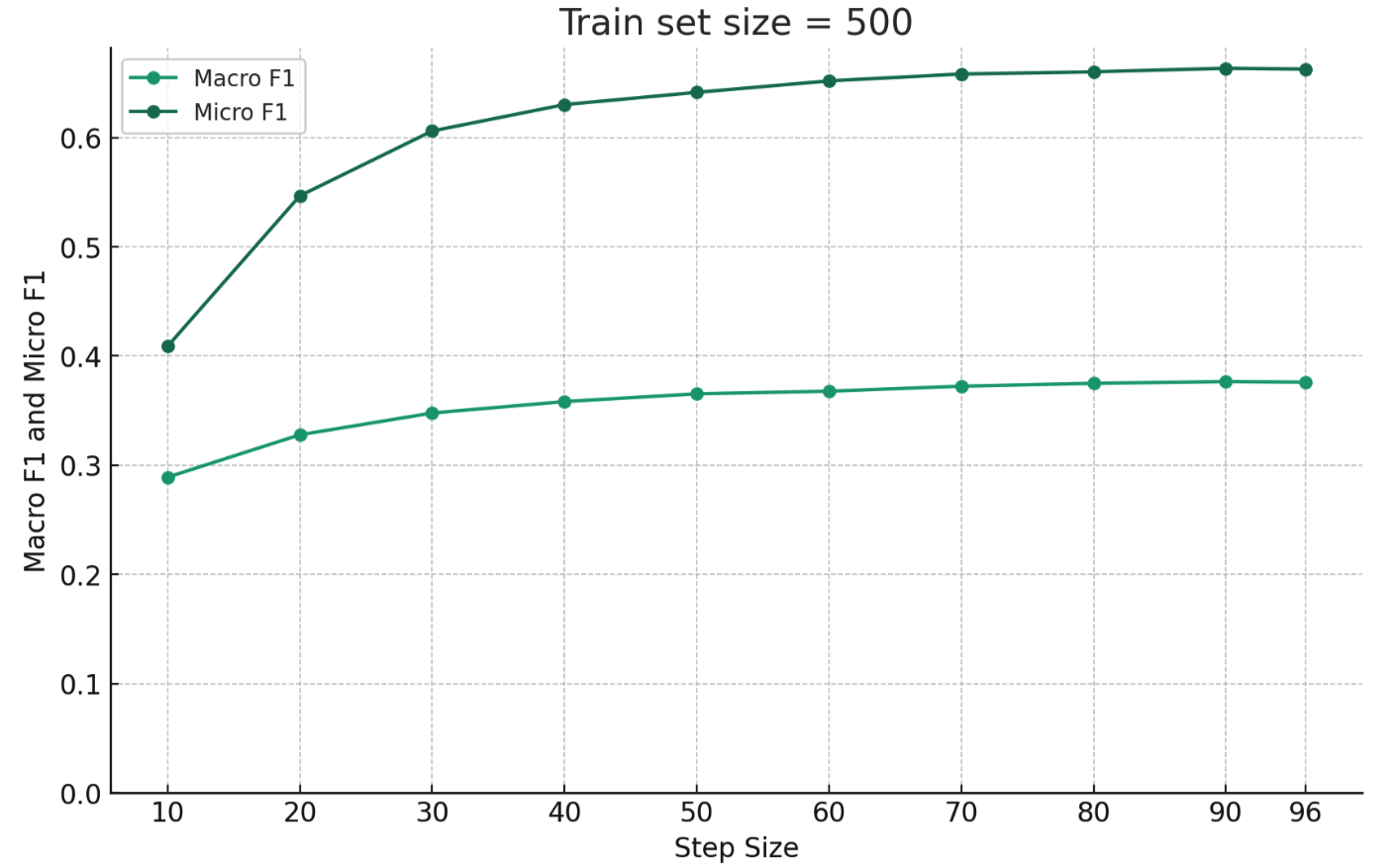}
    \caption{Train set size = 500}
    \label{fig:train500}
  \end{subfigure}

    \centering
  \begin{subfigure}[b]{0.32\textwidth}
    \centering
    \includegraphics[width=\textwidth]{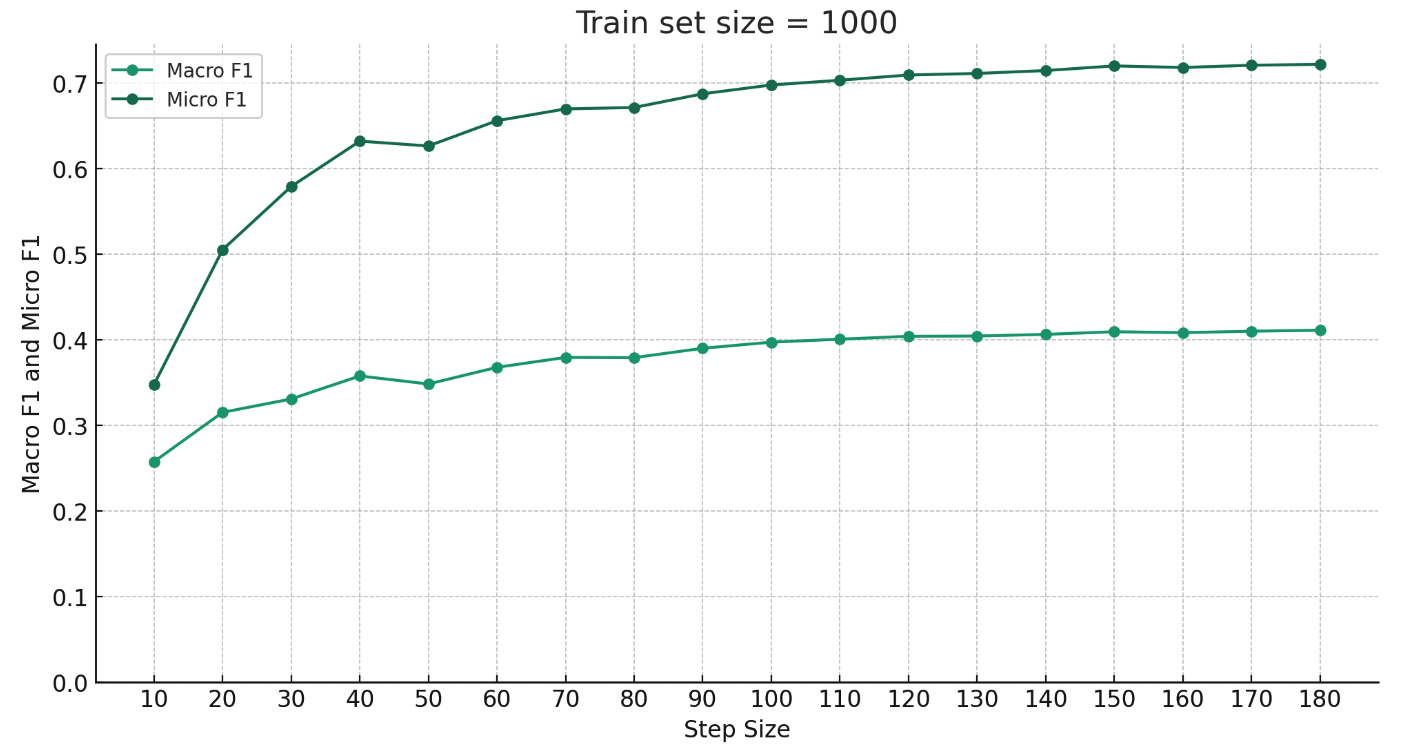}
    \caption{Train set size = 1000}
    \label{fig:train1000}
  \end{subfigure}
  \hfill
  \begin{subfigure}[b]{0.32\textwidth}
    \centering
    \includegraphics[width=\textwidth]{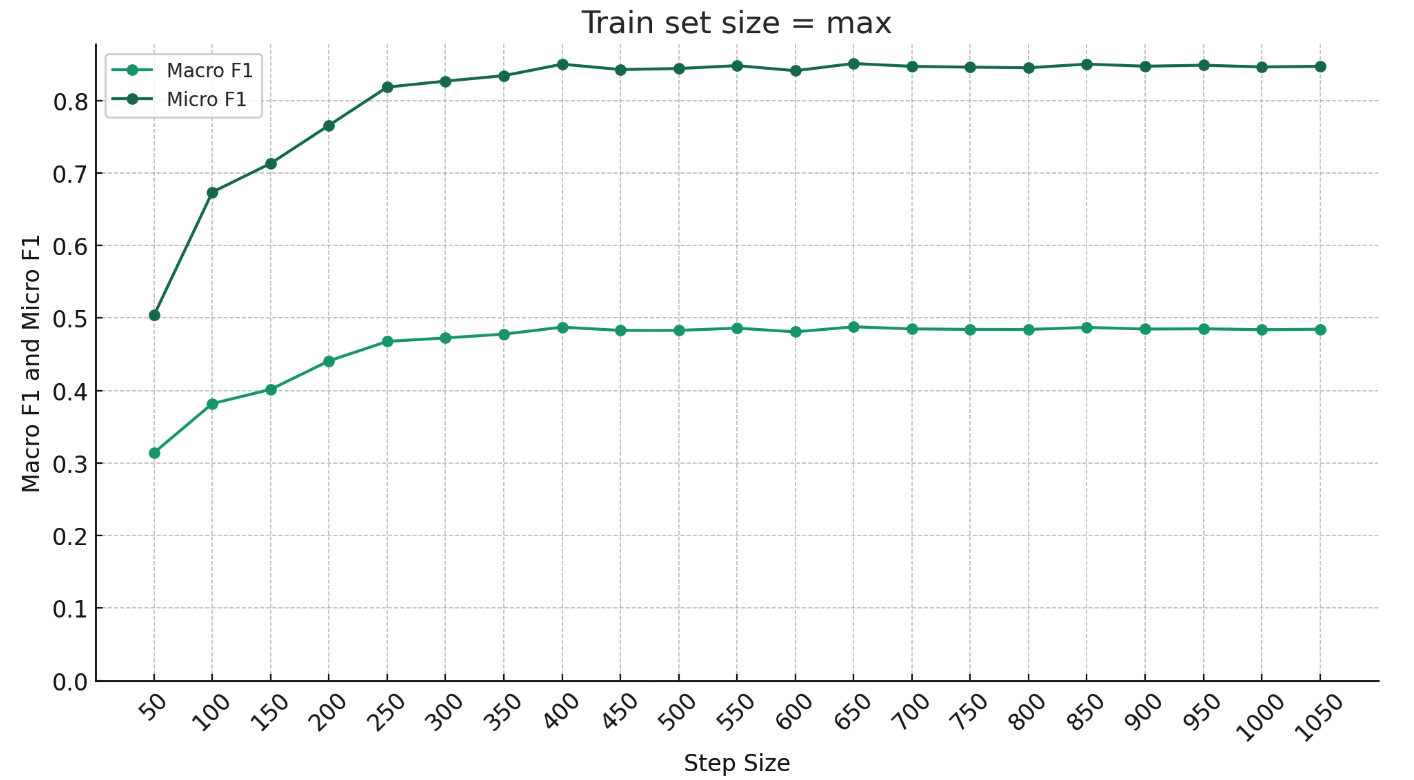}
    \caption{Train set size = max (80\%)}
    \label{fig:trainmax}
  \end{subfigure}
  \hfill
  \begin{subfigure}[b]{0.32\textwidth}
    \centering
    \includegraphics[width=\textwidth]{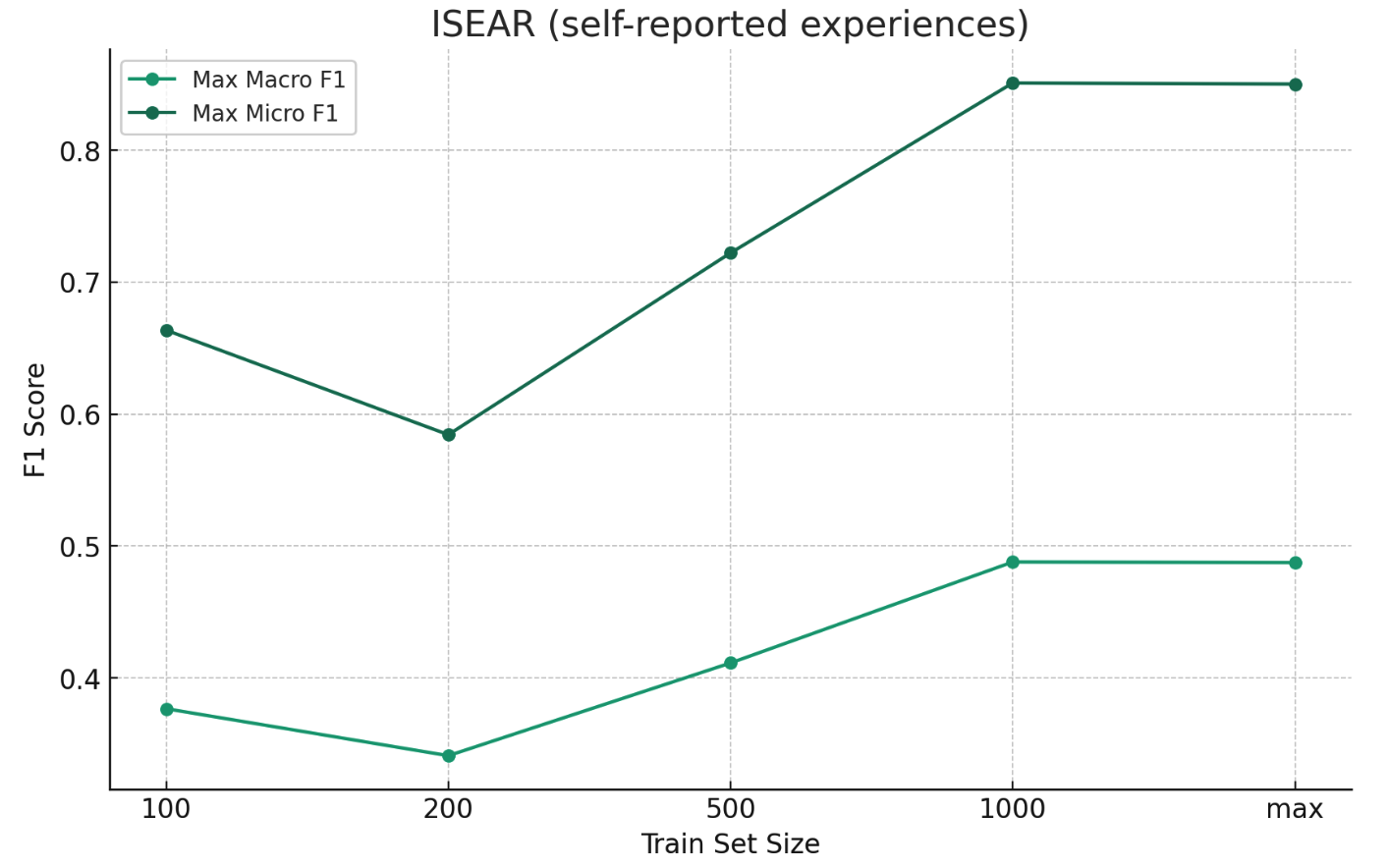}
    \caption{Transfer Learning}
    \label{fig:allts}
  \end{subfigure}

  \caption{Fine-tuning Losses on Three Taxonomies}
  \label{fig:TransferLearning}
\end{figure*}

From Figure \ref{fig:allts} it is easy to find that we reach nearly 90\% in Micro F1 Score while the paper only reached around 60\%. This is a very noticeable improvement in our reproduction of the paper.

\subsection{Analysis and Insights}
For the first set of experiments where we fine-tuned BERT model on three taxonomies, if we look into detail of F1 score for each emotion category, we almost achieved comparable or even better results than the original paper. Especially for the "grief" category in the original taxonomy where the precision, recall and F1 are all 0 in the original paper, we achieved precision of 0.43, recall of 0.50 and F1 score of 0.46 respectively. This is probably due to more training epochs that we adopted.

However, there are still much space for improvement. Firstly, the original GoEmotions dataset is still severely imbalanced. For instance, there are more than 15k data points for the emotion category "admiration", but less than 1k data points for the emotion category "relief". As a result, we achieved 0.67 F1 score on "admiration" category while only 0.32 F1 score on the "relief" category. We suggest that some data augmentation methods can be applied on the minority classes to further improve the model performance~\cite{su2022yolo}. Moreover, our model performance improvements mainly benefit from using more training epochs. So we believe that executing thorough and proper hyperparameter tuning process can help yield better performance as well.

\subsection{Is Fine-tuned RoBERTa A Stronger Baseline?}
In comparison, we propose to fine-tune a RoBERTa model \cite{liu2019roberta} on the GoEmotions training set as the new baseline to replace the BERT baseline proposed in the original paper \cite{demszky-etal-2020-goemotions}. Since RoBERTa optimizes the set of pre-training tasks, the pre-training configuration, and the pre-training dataset \cite{liu2019roberta}, it is expected to outperform the performance of the BERT baseline on GoEmotions in our use case.

We keep exactly the same hyperparameter configuration for RoBERTa as how BERT was trained in the original paper: batch\_size of 16, learning\_rate of 5e-5 and no learning rate scheduler. We use the AdamW optimizer \cite{loshchilov2017decoupled} and cross-entropy as the loss function. Same as BERT baseline, we trained RoBERTa for 4 epochs on the training set to prevent overfitting and the performance comparison with BERT baseline is shown in Table \ref{tab:roberta-original}.

\begin{table}[tp]
\centering
\begin{tabular}{cccc}
\hline
\textbf{Model} & \textbf{Precision} & \textbf{Recall} & \textbf{F1}\\
\hline
BERT & 0.49 & \textbf{0.50} & \textbf{0.49} \\
RoBERTa & \textbf{0.53} & 0.45 & 0.47 \\
\hline
\end{tabular}
\caption{\label{tab:roberta-original}
Macro-averaged Metrics between BERT and RoBERTa Baseline
}
\end{table}

We uncover that the BERT baseline outperforms the RoBERTa baseline in terms of the macro-averaged F1 score after being trained with the same configuration, which is contradict to our initial hypothesis. One possible explanation is that machine learning models are task-sensitive, which means even if RoBERTa generally performs better on a wide range of NLP tasks, there might be specific characteristics of the GoEmotions classification task that make BERT more suitable. For example, the GoEmotions dataset could be more closely aligned with the data distribution which BERT was pre-trained on. In this case, BERT might perform better than RoBERTa on this specific emotion classification task.

Given the experiment results, we disprove the $1^{st}$ hypothesis stated in section \ref{limitations}. And the $2^{nd}$ and $3^{rd}$ experiment design needs to be revised correspondingly:
\begin{enumerate}
    \setcounter{enumi}{1}
    \item (revised) apply three data augmentation techniques on original training dataset before fine-tuning \textbf{BERT}, comparing the variance in performance
    \item (revised) fine-tune \textbf{BERT} on CARER dataset first and then fine-tune the trained model on augmented GoEmotions training dataset, comparing the variance in performance
\end{enumerate}

\begin{figure*}[h!]
  \centering
  \begin{subfigure}[b]{0.32\textwidth}
    \centering
    \includegraphics[width=\textwidth]{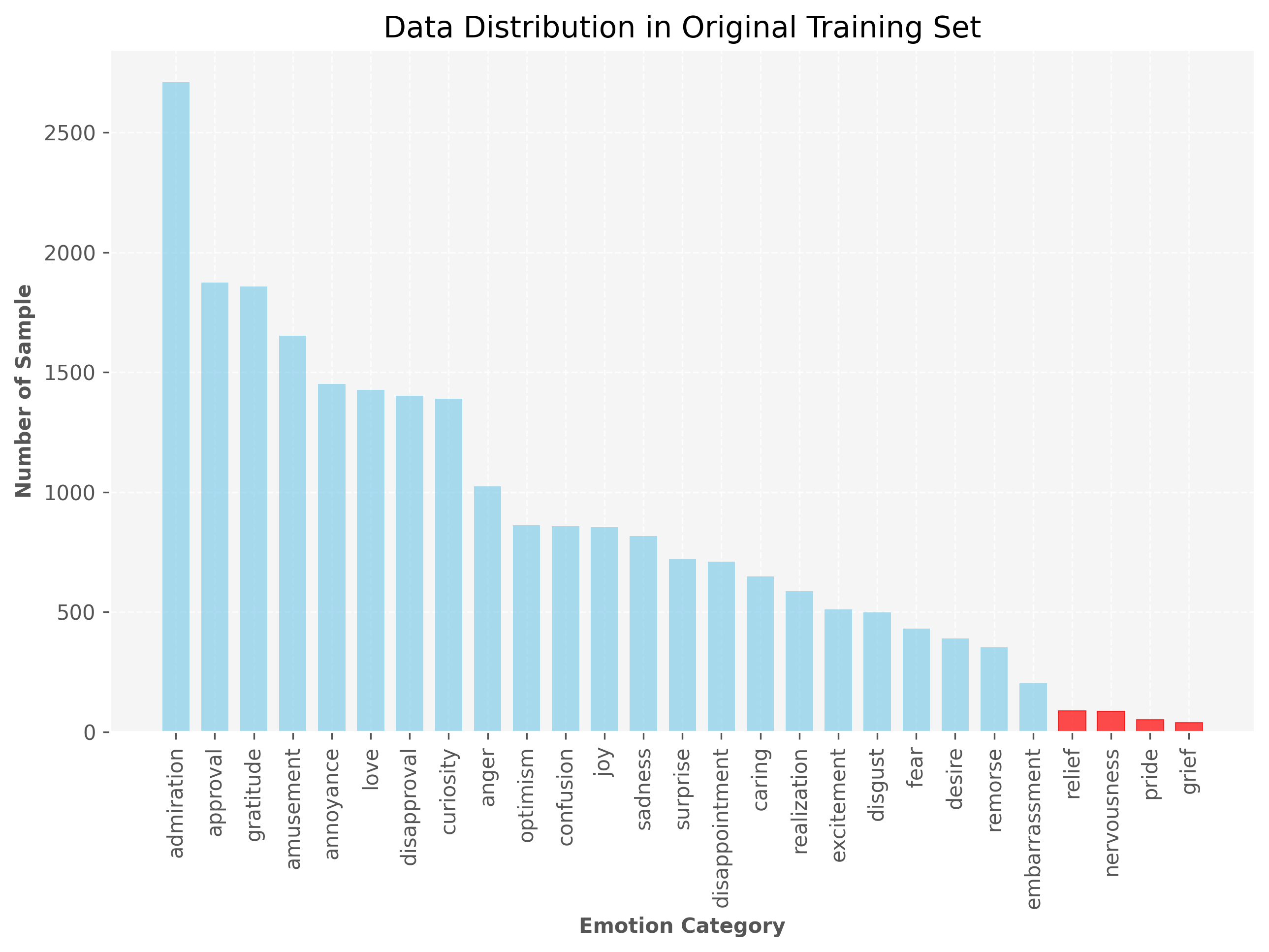}
    \caption{Original Training Set}
    \label{fig:hist_original}
  \end{subfigure}
  \hfill
  \begin{subfigure}[b]{0.32\textwidth}
    \centering
    \includegraphics[width=\textwidth]{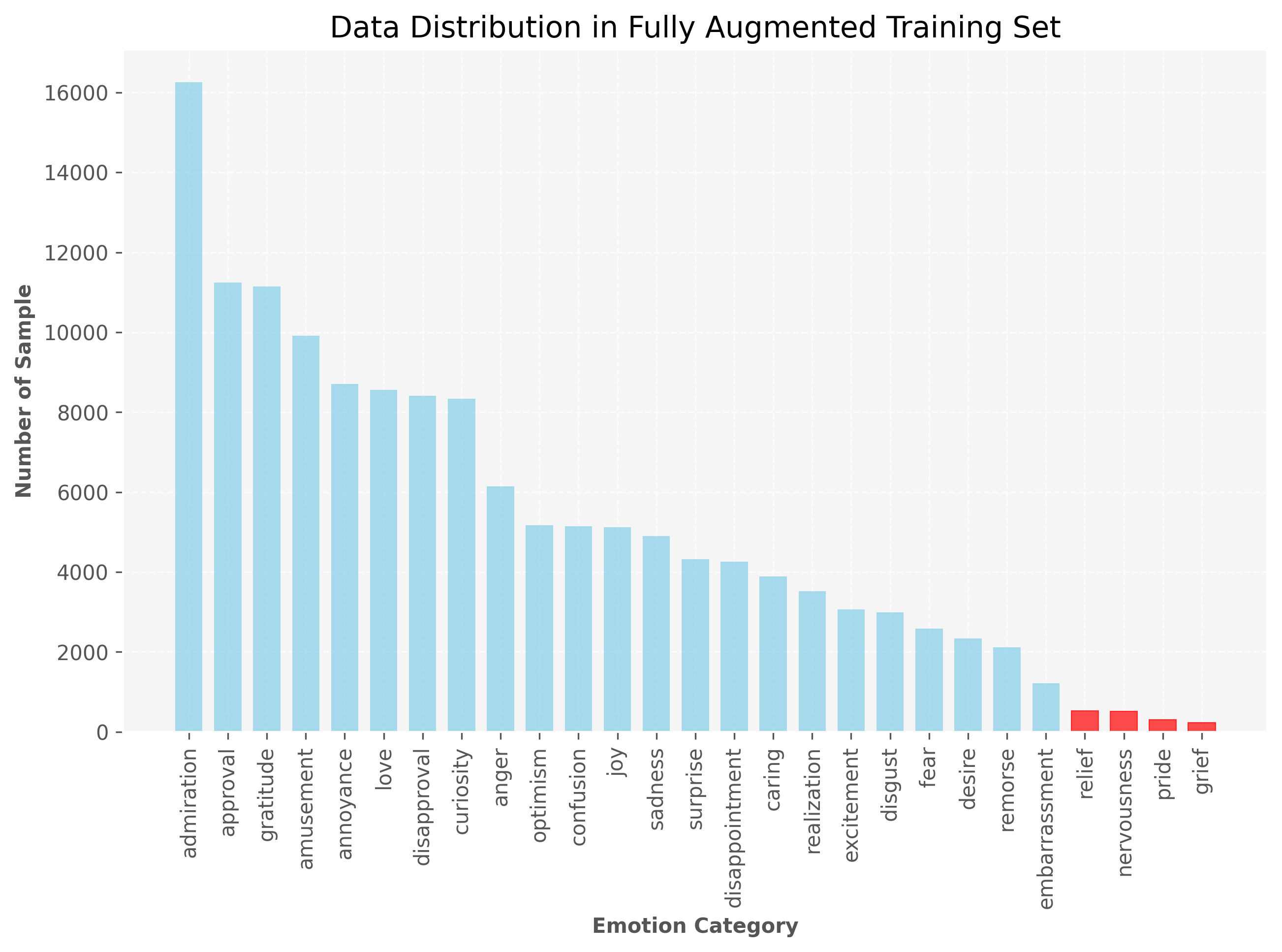}
    \caption{Fully Augmented Training Set}
    \label{fig:hist_fullAug}
  \end{subfigure}
  \hfill
  \begin{subfigure}[b]{0.32\textwidth}
    \centering
    \includegraphics[width=\textwidth]{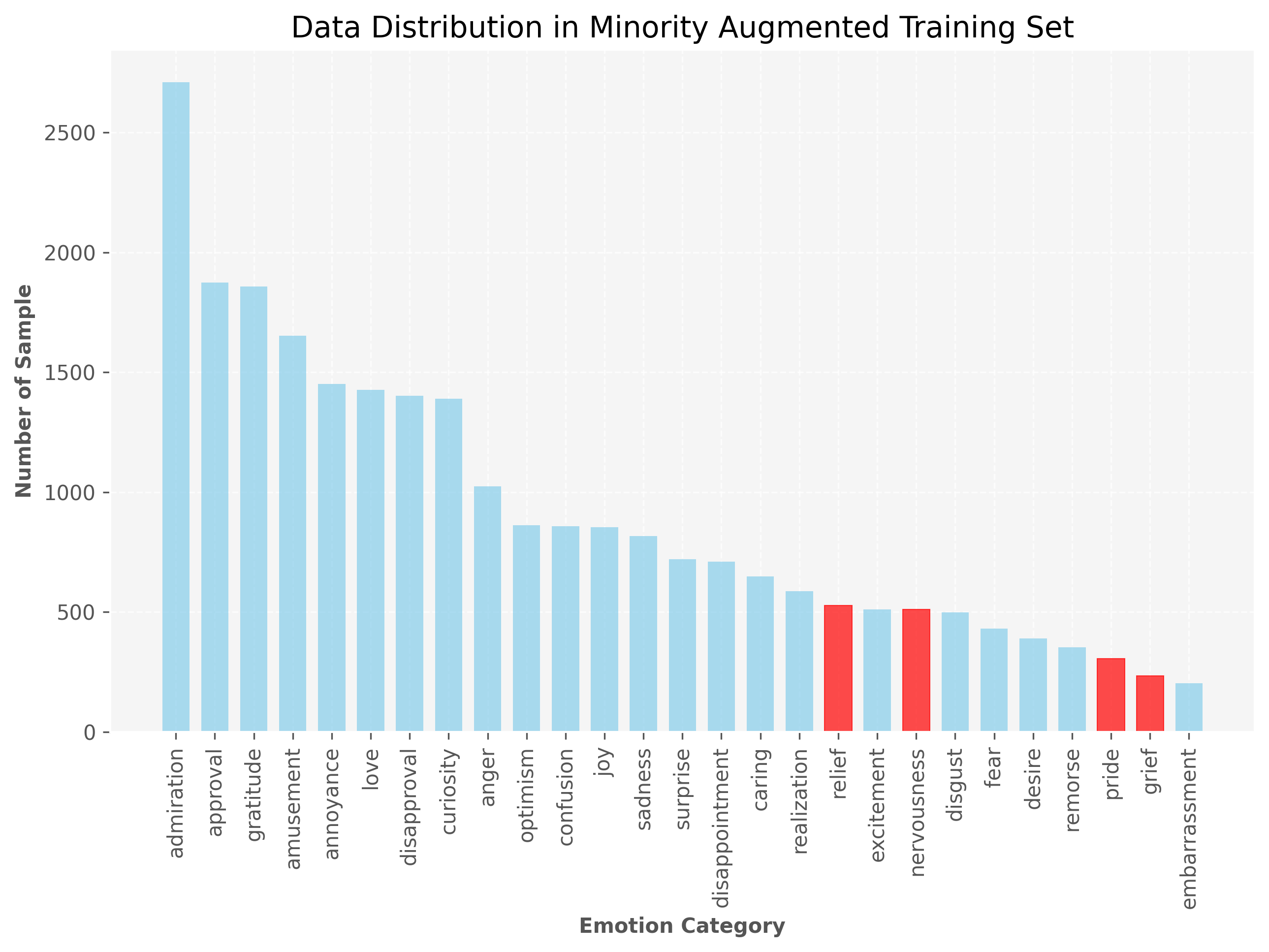}
    \caption{Minority Augmented Training Set}
    \label{fig:hist_minorityAug}
  \end{subfigure}
  
  \caption{Data Distribution across Datasets}
  \label{fig:hist}
\end{figure*}

\subsection{Data Augmentation}
In the revised second set of experiments, we apply three data augmentation techniques on the original GoEmotions training set which contains 43410 data points. The data distribution of emotion categories in the original training set can be found at Figure \ref{fig:hist_original}. Since the "neutral" class dominants the dataset, we remove it from the visualization plot and only keep the rest 27 emotion categories. It can be easily seen that the distribution of original training set is imbalanced. Some categories such as "admiration" has more than 25k examples while some minority classes like "grief" only has 39 examples.

To mitigate the unevenness of the training data distribution, we applied 3 different data augmentation methods on each data point in the training set.

\paragraph{Duplication Data Augmentation (DDA)}
This straightforward method proposed by Wei et al. \cite{wei2019eda} has proven to be notably effective. DDA integrates various manipulations for the words present in the initial document. In our use case, for each sentence within the training data, DDA selects and executes one of the following operations with certain predefined probability:
\begin{itemize}
  \item Synonym Replacement: random words are replaced by random synonyms from a dictionary e.g. WordNet \cite{miller1995wordnet}
  \item Random Swap: randomly chosen words are swapped
  \item Random Deletion: random words are removed
\end{itemize}
This is a quite native approach since we are changing random words in the sentence, which means those words that have higher weights in the classification task may be affected.

\paragraph{Traditional Language Model Embeddings}
To address the drawback in DDA approach, a more sophisticated way to altering the original text without hurting the meaning and context of the sentence is to use BERT embeddings \cite{devlin-etal-2019-bert}. Essentially, the word \textit{w} in the original text is substituted with words forecasted by a traditional Language Model (LM) like BERT, considering the surrounding context of \textit{w} in the original text. In our use case, a new word will be injected to random position according to contextual word embeddings or an original word will be replaced by a new word according to contextual embedding distances with predefined probability.

\paragraph{BART Paraphraser ProtAugment}
The third method ProtAugment from \cite{dopierre2021protaugment} shifts the focus from individual words to rephrasing the entire original text. It uses BART, a model that "merges Bidirectional and Auto-Regressive Transformers" \cite{lewis2019bart}, for creating paraphrases. The authors of ProtAugment tuned BART specifically for paraphrase generation, employing multiple datasets for this purpose. Additionally, they employed techniques like Diverse Beam Search \cite{vijayakumar2018diverse} and Back Translation \cite{mallinson2017paraphrasing} to help producing diverse outputs.

We augmented each training data point to generate 5 new data points per method, resulting in three fully augmented training set. Each fully augmented training set is 5 times larger than the original training set. We randomly picked one fully augmented training set (the one used ProtAugment method) to fine-tune a BERT model on it with the same training configuration as our baseline, the results are shown in Table \ref{tab:fully_aug_train}.

\begin{table}[tp]
\centering
\begin{tabular}{cccc}
\hline
\textbf{Training Set} & \textbf{Precision} & \textbf{Recall} & \textbf{F1}\\
\hline
Original & \textbf{0.49} & \textbf{0.50} & \textbf{0.49} \\
Fully Augmented & \textbf{0.49} & 0.41 & 0.44 \\
\hline
\end{tabular}
\caption{\label{tab:fully_aug_train}
Macro-averaged Metrics on Original and Fully Augmented Training Set
}
\end{table}

The experiment results show that the model performance on fully augmented dataset is not as good as our BERT baseline on the original training set. However, this is not surprising. Referring to the data distribution of fully augmented training set in Figure \ref{fig:hist_fullAug}, we find that the shape of the distribution nearly stays the same but the scale of the distribution is expanded, which exacerbates the unevenness. This can also be proved by a dramatic increment in terms of standard deviation after fully augmented the dataset stated in Table \ref{tab:eveness}. Therefore, the model will learn better on the majority classes in the fully augmented dataset, but learn even worse on those minority classes than our baseline.

\begin{table}[tp]
\centering
\begin{tabular}{cc}
\hline
\textbf{Training Set} & \textbf{Std}\\
\hline
Original & 2350.36 \\
Full Augmented & 14102.14 \\
Minority Categories Augmented & \textbf{2303.25} \\
\hline
\end{tabular}
\caption{\label{tab:eveness}
Data Distribution Evenness across Datasets
}
\end{table}

One mitigation strategy is to only augment four minority classes in the original training set, namely "grief", "pride", "nervousness", and "relief". Similarly, we augmented each training data point in minority categories to generate 5 new data points per method, resulting in 3 different minority augmented training set. The new data distribution of the minority augmented training set can be found in Figure \ref{fig:hist_minorityAug} with four minority classes highlighted in red. We also find that the standard deviation of the minority augmented training set distribution is much lower than that of the original training set (see Table \ref{tab:eveness}), which means the original data unevenness is successfully mitigated.

Then our team tuned BERT model on three minority augmented dataset using the same training configuration as the baseline, used the same original testing set to do prediction, and reported the results in Table \ref{tab:minority_aug_train}.

\begin{table}[tp]
\centering
\begin{tabular}{cccc}
\hline
\textbf{Training Set} & \textbf{Precision} & \textbf{Recall} & \textbf{F1}\\
\hline
Original & 0.490 & 0.500 & 0.490 \\
DDA & 0.500 & \textbf{0.537} & 0.511 \\
BERT Embed & 0.518 & 0.528 & 0.511 \\
ProtAug & \textbf{0.519} & 0.533 & \textbf{0.517} \\
\hline
\end{tabular}
\caption{\label{tab:minority_aug_train}
Macro-averaged Metrics on Three Minority Augmented Training Sets
}
\end{table}

These results show that all three augmentation methods on minority emotion categories can improve the BERT classification performance, which aligned with our expectation stated in the $2^{nd}$ hypothesis. Among them, the model tuned on the ProtAugment training set gives the best performance in terms of macro-averaged F1 score, we also include the detailed category-wise performance for this model in Table \ref{tab:best_prot}.

\begin{table}[tp]
\centering
\begin{tabular}{lccc}
\hline
\textbf{Emotion} & \textbf{Precision} & \textbf{Recall} & \textbf{F1}\\
\hline
admiration & 0.63 & 0.74 & 0.68 \\
amusement & 0.72 & 0.94 & 0.82 \\
anger & 0.48 & 0.50 & 0.49 \\
annoyance & 0.36 & 0.29 & 0.32 \\
approval & 0.53 & 0.35 & 0.42 \\
caring & 0.40 & 0.42 & 0.41 \\
confusion & 0.46 & 0.43 & 0.44 \\
curiosity & 0.46 & 0.77 & 0.57 \\
desire & 0.60 & 0.41 & 0.49 \\
disappointment & 0.31 & 0.23 & 0.26 \\
disapproval & 0.40 & 0.33 & 0.36 \\
disgust & 0.51 & 0.48 & 0.50 \\
embarrassment & 0.67 & 0.43 & 0.52 \\
excitement & 0.51 & 0.36 & 0.42 \\
fear & 0.58 & 0.78 & 0.66 \\
gratitude & 0.94 & 0.90 & 0.92 \\
grief & 0.20 & 0.33 & 0.25 \\
joy & 0.57 & 0.61 & 0.59 \\
love & 0.71 & 0.89 & 0.79 \\
nervousness & 0.36 & 0.35 & 0.36 \\
optimism & 0.63 & 0.56 & 0.60 \\
pride & 0.50 & 0.38 & 0.43 \\
realization & 0.27 & 0.19 & 0.22 \\
relief & 0.45 & 0.45 & 0.45 \\
remorse & 0.58 & 0.95 & 0.72 \\
sadness & 0.50 & 0.59 & 0.54 \\
surprise & 0.60 & 0.53 & 0.56 \\
neutral & 0.62 & 0.72 & 0.67 \\
\textbf{macro-average} & \textbf{0.52} & \textbf{0.53} & \textbf{0.52} \\
\textbf{std} & \textbf{0.15} & \textbf{0.22} & \textbf{0.17} \\
\hline
\end{tabular}
\caption{\label{tab:best_prot}
Category-wise Performance of BERT Model Trained on ProtAugment Training Set
}
\end{table}

\subsection{Transfer Learning}

In this section, we want to test the impact of introducing Transfer Learning into our experiment settings. This idea is learned from the original paper of GoEmotions \cite{demszky-etal-2020-goemotions}, and we use CARER \cite{saravia-etal-2018-carer}, another Emotion Classification Task Dataset, as additional data source.

\subsubsection{CARER Overview}
The CARER dataset is an Emotion Detection dataset that was collected using noisy labels and annotated via distant supervision. This dataset focuses on 6 primary emotions: \textit{anger, fear, joy, love, sadness, and surprise}. The dataset is specifically tailored for emotion recognition tasks and is used in the context of multi-class and multi-label text classification, as well as in semantic textual similarity tasks. It serves as a significant resource for researchers and practitioners in the field of natural language processing, specifically for those working on emotion recognition from text. Table \ref{tab:Data_carer} shows the label distribution of CARER.
\begin{table}[tp]
\centering
\begin{tabular}{ccc}
\hline
\textbf{Emotions} & \textbf{Amount} & \textbf{Hashtags} \\ \hline
sadness & 214,454 & \#depressed, \#grief \\
joy & 167,027  & \#fun, \#joy \\
fear & 102,460  & \#fear, \#worried \\
anger & 102,289  & \#mad, \#pissed \\
surprise & 46,101  & \#strange, \#surprise \\
trust & 19,222  & \#hope, \#secure \\
disgust & 8,934  & \#awful, \#eww \\
anticipation & 3,975  & \#pumped, \#ready \\
\hline
\end{tabular}
\caption{\label{tab:Data_carer}
Label Distribution of CARER
}
\end{table}
The CARER paper and dataset contribute significantly to the field of affective computing and natural language processing by providing a nuanced approach to emotion recognition in text, reflecting the complex ways in which emotions are expressed and understood in human language.

\begin{table*}[tp]
\centering
\begin{tabular}{lcccc}
\hline
\textbf{Experiment} & \textbf{Accuracy} & \textbf{Precision} & \textbf{Recall} & \textbf{F1 Score} \\ \hline
BERT on CARER (reference) & 93.75\%  & 92.21\% & 91.14\% & 91.58\% \\ 
BERT on original GoEmotions (original paper)& N/A  & 40\% & 63\% & 46\% \\ 
BERT on PROT Augmented GoEmotions & N/A  & 51.9\% & 53.3\% & 51.7\% \\ \hline
CARER-BERT on original GoEmotions & 48.54\%  & 55.66\% & 46.06\% & 47.38\% \\
CARER-BERT on DDA Augmented GoEmotions & 49.19\%  & \textbf{57.97\%} & 46.38\% & 50.03\% \\
CARER-BERT on BERT Augmented GoEmotions & 48.47\%  & 57.69\% & 47.15\% & 50.52\% \\
CARER-BERT on PROT Augmented GoEmotions & \textbf{49.32\%}  & 57.27\% & \textbf{49.18\%} & \textbf{51.83\%} \\ \hline
GPT-4 on original GoEmotions & 34.91\%  & 10.17\% & 16.90\% & 12.54\% \\
\hline
\end{tabular}
\caption{\label{tab:carer_res}
Results for Transfer Learning using CARER, evaluated on the original test set of GoEmotions.
}
\end{table*}

\subsubsection{Experiment Setting} \label{carer_exp}
We pick BERT \cite{devlin-etal-2019-bert} as our base model, which is aligned with our other experiments, and we first finetune the base model on CARER \cite{saravia-etal-2018-carer}. Then we train our model on 4 different Datasets: Oringinal GoEmotions, DDA Augmented GoEmotions, BERT Embedding Augmented GoEmotions, PROT Augmented GoEmotions. Note that all augmented methods are only applied to under-performanced categories.

We fine-tune BERT on CARER using the hyperparameters of: learning rate of 2e-5, epoch of 10, weight decay of 0.01, warm up steps of 500, batch size of 16 for both training and evaluating, and we set the random seed to 11711 to make sure reproducibility. 

We use trainer of HuggingFace \cite{wolf-etal-2020-transformers} as main code frame and migrate our training process to GPU devices using HuggingFace Accelerate \cite{accelerate}.

\subsubsection{Transfer Learning Results}

We show the results of Transfer Learning by adding CARER into the game, in Table \ref{tab:carer_res}. As mentioned in the above section \ref{carer_exp}, we have in total 4 sets of experiments, which are the BERT finetuned on CARER on 4 different datasets. We evaluate the performance using the original GoEmotions test set, and we report the results according to the best epoch among the 10 epochs trained. In all, we report the model predictions on macro average of Accuracy, Precision, Recall, and F1 Score.

The notation of CARER-BERT means using BERT finetuned on CARER as base model.

\subsection{Modern Large Language Models on GoEmotions}

As discussed above as the fifth hypothesis, we aim to test the performance of the modern LLMs on Emotione Detection task, where \"modern LLMs\" refers to LLMs that have billion-level parameter size \cite{zhao2023survey}, differing to those with million-level like BERT tested in the original GoEmotions paper. The experiment settings for testing modern LLMs is that we use a subset of the GoEmotions dataset, 1000 data points, as input to LLMs with the following prompts:

"I want see your performance in emotion detection task. The setting aligns with the GoEmotions paper with title of "GoEmotions: A Dataset for Fine-Grained Emotion Classification". There are in total 28 categories as emotion labels and every data entry can have one or more of the 28 categories. I will give you a number of sentences that may or may not contain certain emotions inside, and your job is to label the input sentences with the given 28 categories like the setting of GoEmotions paper. Please organize the labels into a string of words with comma as separator. The output should be a csv table with 3 columns where every row contains the id, the given sentence, and your label."

We use the above text as prompt and we call OpenAI's API for ChatGPT-4 using 30 sentences as a batch input in every call. We then collect their response and calculate the accuracy. The calculation on metrics is different than what we did to BERT and RoBERTa and we will address this in the Analysis section. Note that we particularly want to test their zero-shot performance on emotion detection task thus there is no training input. We use GPT-4 as the modern LLM and we skip including Llama in the scope as it cannot tell the 28 categories correctly and it can only tell emotions from positive, negative, and neutral.

\section{Error Analysis}
\subsection{Data Augmentation Analysis}

\begin{table}[tp]
\centering
\begin{tabular}{lcc}
\hline
\textbf{Emotion} & \textbf{F1-Original} & \textbf{F1-ProtAug}\\
\hline
admiration & 0.65 & \textbf{0.68} \\
amusement & 0.80 & \textbf{0.82} \\
anger & 0.47 & \textbf{0.49} \\
annoyance & \textbf{0.34} & 0.32 \\
approval & 0.36 & \textbf{0.42} \\
caring & 0.39 & \textbf{0.41} \\
confusion & 0.37 & \textbf{0.44} \\
curiosity & 0.54 & \textbf{0.57} \\
desire & \textbf{0.49} & \textbf{0.49} \\
disappointment & \textbf{0.28} & 0.26 \\
disapproval & \textbf{0.39} & 0.36 \\
disgust & 0.45 & \textbf{0.50} \\
embarrassment & 0.43 & \textbf{0.52} \\
excitement & 0.34 & \textbf{0.42} \\
fear & 0.60 & \textbf{0.66} \\
gratitude & 0.86 & \textbf{0.92} \\
grief & 0.00 & \textbf{0.25} \\
joy & 0.51 & \textbf{0.59} \\
love & 0.78 & \textbf{0.79} \\
nervousness & 0.35 & \textbf{0.36} \\
optimism & 0.51 & \textbf{0.60} \\
pride & 0.36 & \textbf{0.43} \\
realization & 0.21 & \textbf{0.22} \\
relief & 0.15 & \textbf{0.45} \\
remorse & 0.66 & \textbf{0.72} \\
sadness & 0.49 & \textbf{0.54} \\
surprise & 0.50 & \textbf{0.56} \\
neutral & \textbf{0.68} & 0.67 \\
\textbf{macro-average} & 0.46 & \textbf{0.52} \\
\textbf{std} & 0.19 & \textbf{0.17} \\
\hline
\end{tabular}
\caption{\label{tab:compare_f1}
Compare F1 Score between BERT Baseline and BERT ProtAugment
}
\end{table}

To conduct a comprehensive error analysis on the effectiveness of the data augmentation, we list the category-wise F1 scores of BERT tuned on the original training set and BERT tuned on the ProtAugment training set in Table \ref{tab:compare_f1}.

On the whole, the macro-average F1 score has improved from 0.46 to 0.52 with the use of the augmented training set, indicating that the augmentation process has generally benefited the model's ability to classify emotions across all categories. Moreover, the standard deviation of the F1 scores across emotions has decreased from 0.19 to 0.17, which suggests that the augmentation has not only improved the overall performance but also forced the model to learn more consistently across different emotion categories.

If we look into single category granularity, 23 out of 28 classes see improvements in F1 score. Especially for four augmented categories "grief", "pride", "nervousness", and "relief", we can see significant increment in their classification performance, indicating that the augmented training set provided sufficient examples in these four categories for the model to learn from, where previously it struggled to recognize these emotions. For the rest 19 categories with no augmentation but improved, we may conclude that the model can also benefit indirectly from the augmentation of other classes and directly benefit from the evenness of the data distribution. A few categories such as "annoyance", "disappointment", "desire", "disapproval", and "neutral" see a decrease in performance. This may indicate that the augmented data introduced some noise or confusion for these specific emotions, or there could be an issue of class imbalance that the augmentation didn't address.

Based on the analysis, future work may involve further augmentation for underperforming categories or making the training set more balanced to ensure that the model can generalize well across all emotion categories.

\subsection{Transfer Learning Analysis}
From Table \ref{tab:carer_res}, we can easily gain several insights listed below:
\begin{itemize}
    \item CARER's incorporation demonstrably improves performance. As evidenced by the 2nd and 3rd rows of the table, CARER-BERT achieves a statistically significant 0.9\% F1 score improvement over vanilla BERT.
    \item Data augmentation strategies further boost gains. CARER-BERT augmented with data for underperforming categories exhibits a remarkable 4\% F1 score increase, highlighting the efficacy of this targeted approach.
    \item PROT augmentation emerges as the most effective strategy. Among the three augmentation methods employed, PROT consistently delivers the best results, similar to the last section.
    \item On PROT augmented dataset, CARER-BERT works better than BERT. The 3rd and last rows showcase CARER-BERT's, though slightly, better performance on PROT-augmented data, demonstrating Transfer Learning's positive impact.
\end{itemize}

Note that, to our analysis, F1 score is the main metric that we use to compare the models' performance because it reflects the correctness of model predictions most correctly and comprehensively. Overall, we have successfully added CARER to our base models and it proves to be helpful to our model's classification performance.

\subsection{Modern LLMs on GoEmotions Analysis}

As shown in the table 12, in zero-shot setting, comparing to fine-tuned BERT and RoBERTa, the performance of GPT-4 is much lower in accuracy, precision, recall and F1-score. Note that here we keep the same way of calculating metrics of using macro average on every emotion class. We summarize the reasons of such low performance as followings:

\paragraph{Hallucination}. The problem of hallucination is well-known to be associated with generative models and LLMs \cite{jing2024large}, and we found in our case, GPT-4 will classify a sentence with a emotion label that is not included in the 28 categories. Here is an example: the given sentence is "10 life hacks you didn't know you needed in your life." and GPT-4 put "curiosity" and "informative" as emotion labels to it, where informative is not one of the designated categories. Similar problems happen to quite a few data points. In fact, to our test of 1000 data points of GoEmotions dataset, there are in total 89 examples with emotion labels that are made up by GPT-4 and not included in the 28 categories in reality. This is to say, by using 30 data points as batch input, there are almost one made-up label in every 10 given input. But we have emphasize that 28-class classification task is much harder than 3 classes of positive, negative, and neutral as labels in other common emotion detection task.

\paragraph{Over-labelling}. We found that almost every predicted sentence have at least 2 labels in GPT-4's response whereas multi-label data points are much less common in the original dataset of GoEmotions. We calculate the number of the examples that have more predicted labels than the true label, and we found up to 812 out of 1000 data points. The majority of such examples exhibit the problem of correctly predicting the true labels while adding a new emotion label that is similar or close in semantic meaning of the predicted ones. This would decrease the result in precision, recall, and F1 score.

\paragraph{Over-interpretation}. Often the case is GPT-4 will give a response that overly interprete the meaning of the sentence. For example, the given sentence is "Ask him out for a drink.", and in normal occasion this is simply a proposal to ask somebody out. The true label for this sentence in the dataset is "neutral" whereas the response from GPT-4 is "desire" and "optimism". This is clearly over-interpretation of the meaning of the sentence.

In summary, we are the first to apply modern LLMs to GoEmotions dataset, and we observed 3 challenges when employing GPT-4 for emotion classification tasks, especially when benchmarked against fine-tuned models like BERT and RoBERTa.

\section{Conclusion and Future Work}

In this project, the study embarked on an exploratory journey to enhance the classification performance on the fine-grained GoEmotions dataset. Through meticulous experimentation, we have validated the efficacy of data augmentation and transfer learning as viable strategies to improve emotion detection in text. Notably, our findings indicate that the integration of the CARER dataset via transfer learning resulted in a measurable performance boost, as demonstrated by the improved F1 scores across multiple configurations.

Data augmentation, particularly when applied to under-performing categories, has proven to be a potent tool in balancing the dataset \cite{dong2021variational} and enhancing model accuracy. Among the augmentation techniques, PROT augmented dataset has emerged as the superior strategy, outperforming other methods in improving the classification results of our models.

On top of that, our study also highlights the complexity and sensitivity of machine learning tasks to their respective datasets. The initially hypothesized superiority of the RoBERTa model was not realized, underscoring the need for task-specific model selection and configuration. Furthermore, we observed that while data augmentation can indeed propel performance forward, it requires a nuanced application to prevent exacerbating existing data imbalances. Also, we identified three primary issues contributing to GPT-4's inferior performance metrics: hallucination, over-labelling, and over-interpretation when applying GPT-4 to GoEmotions dataset.

Our research contributes to the broader understanding of emotion detection in NLP, offering insights that extend beyond the confines of our dataset.

Future work, on the one hand, could explore the application of these strategies to other NLP tasks and datasets, potentially contributing to the development of more robust, accurate, and fair machine learning models. On the other hand, we found that there are not very many of interests received in the emotion detection domain comparing to the current trend of Large Language Models (LLMs). Thus we think contributing a survey that collectively introduces and analyses different methods' performance on various datasets in this domain would be of great help to the researchers, and we are drawn to the possibility of publishing a survey paper.

\bibliographystyle{apalike}
\bibliography{custom}

\end{document}